\documentclass[sigconf,authorversion]{acmart}
\AtBeginDocument{%
  \providecommand\BibTeX{{%
    \normalfont B\kern-0.5em{\scshape i\kern-0.25em b}\kern-0.8em\TeX}}}

\setcopyright{acmcopyright}
\copyrightyear{2020}
\acmDOI{10.1145/3379156.3391375}
\acmISBN{ 978-1-4503-7134-6/20/06$15.00}

\begin{document}
\title{Privacy-Preserving Eye Videos using Rubber Sheet Model}

\author{Aayush K. Chaudhary}
\email{akc5959@rit.edu}
\affiliation{%
  \institution{Rochester Institute of Technology}
  \city{Rochester}
  \state{NY}
  \country{USA}}
%\email{akc5959@rit.edu}

\author{Jeff B. Pelz}
\email{pelz@cis.rit.edu}
\affiliation{%
  \institution{Rochester Institute of Technology}
  \city{Rochester}
  \state{NY}
  \country{USA}}

\renewcommand{\shortauthors}{Aayush K. Chaudhary \& Jeff B. Pelz}
%
% The abstract is a short summary of the work to be presented in the article.
\begin{abstract}
Video-based eye trackers estimate gaze based on eye images/videos. As security and privacy concerns loom over technological advancements, tackling such challenges is crucial. We present a new approach to handle privacy issues in eye videos by replacing the current identifiable iris texture with a different iris template in the video capture pipeline based on the Rubber Sheet Model. We extend to image blending and median-value representations to demonstrate that videos can be manipulated without significantly degrading segmentation and pupil detection accuracy.
\end{abstract}
%\newline

%\ccsdesc[300]{Artificial intelligence~ Computer vision}
%\ccsdesc{Computer Vision~ Computer Vision problems;}

% Keywords. The author(s) should pick words that accurately describe the work being
% presented. Separate the keywords with commas.

%
% A "teaser" image appears between the author and affiliation information and the body 
% of the document, and typically spans the page. 
%
% This command processes the author and affiliation and title information and builds
% the first part of the formatted document.
\maketitle
%Resetting passwords is easy. Changing your eyes is hard.

\begin{figure}[h]
\begin{center}
\includegraphics[width=0.8\linewidth]{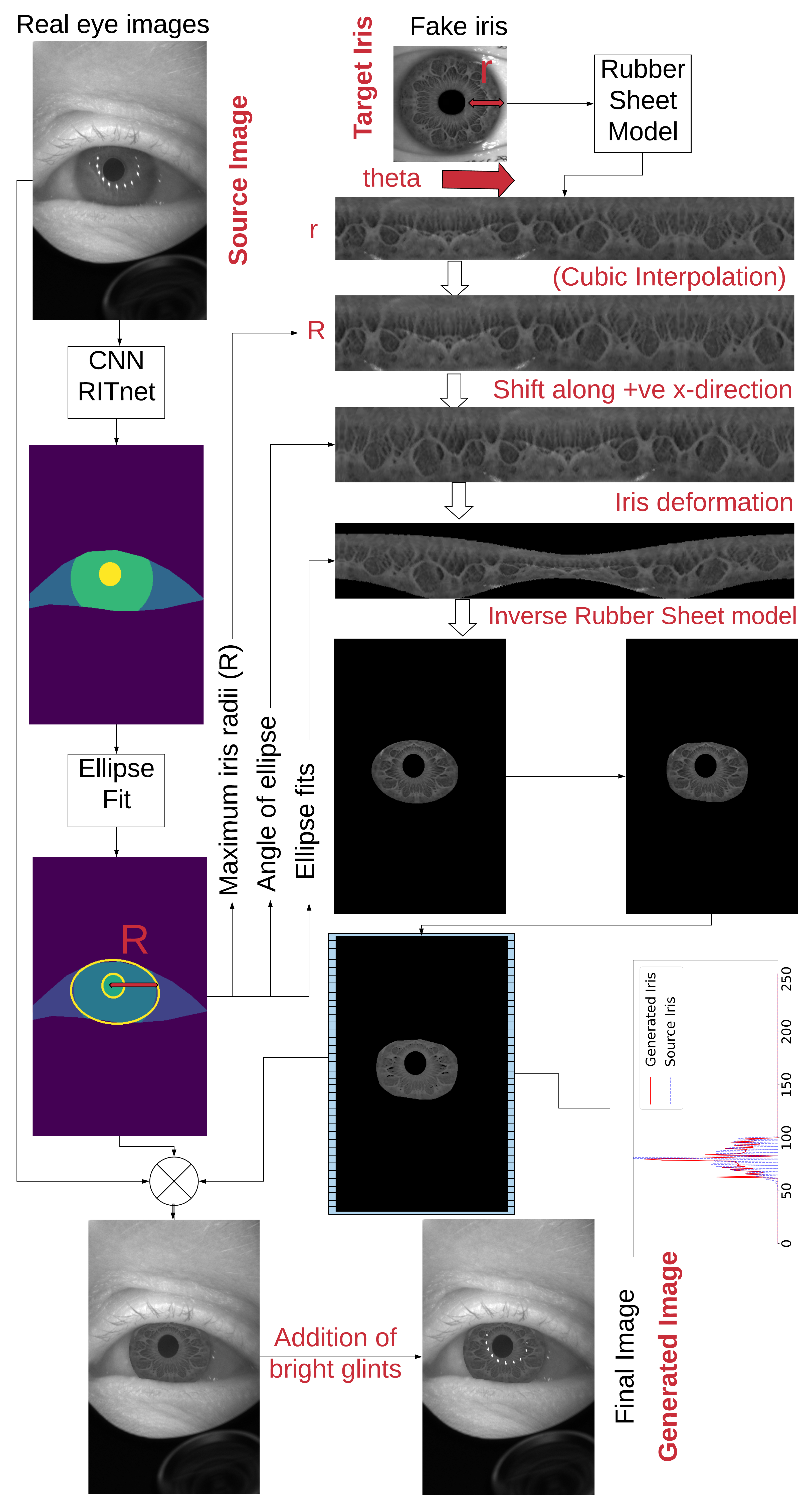}
\end{center}
\caption{Basic flow of proposed method (Section~\ref{methods}). The source image \cite{garbin2019openeds} is passed to a CNN to annotate the eye regions. The target iris undergoes a rubber sheet model transformation \cite{daugman2009iris} followed by other transformations based on the information from the source image. This results in a generated iris similar to the source iris shape, which is then mixed with the source image to get the final image after the glints are replaced.}
\label{fig:block}
\end{figure}

\section{Introduction}
Due to advances in the fields of augmented and virtual reality, the use of eye-tracking is being rapidly extended beyond traditional research, medical diagnosis, and behavioral studies. As the technology extends to the general public, the essential concern of data privacy needs to be addressed.  Eye data is valuable privacy-sensitive information that provides insight about human behavior, private life, health data, biometric signatures, etc. \cite{steil2019privaceye,steil2019privacy, liu2019differential,john2019eyeveil,bozkir2019privacy}. According to a survey conducted by \citet{steil2019privacy}, people are interested in sharing eye-tracking data if it helps them in medical diagnosis or in improving their user experience but not for use in personal or behavioral study. This demands proper privacy regulations and changes in data capture mechanisms to protect user's privacy \cite{liu2019differential}.

Recently \cite{steil2019privaceye,steil2019privacy, liu2019differential,john2019eyeveil,bozkir2019privacy} have proposed solutions for eye-tracking data privacy.  Some of these efforts \cite{steil2019privacy, liu2019differential} are related to preserving privacy in gaze data in addition to the eye images, as it also provides important information regarding the individual's attention, cognitive ability, health, and emotions. The proposed solutions have been to limit the ability to identify individuals' data by aggregating data of multiple participants into a statistical database/representations (differential privacy) \cite{steil2019privacy, liu2019differential}.  Similarly, \citet{steil2019privaceye} proposed a solution to capture the eye tracker's first-person video, based on scene image and eye movements.

To our knowledge, only \cite{john2019eyeveil,bozkir2019privacy} have considered the possibility of identifying individuals based on eye videos. Eye videos contain information about the pupil, iris, sclera, eye corners, eyelids, and face in addition to the eye movements. Among these, the iris contains the most important biometric data, as iris patterns are unique among individuals \cite{daugman2003importance}. \citet{john2019eyeveil} proposed a technique to improve privacy by defocusing the eye image (optically or through digital blurring) while retaining satisfactory accuracy in gaze estimation. Both \citet{phillips2011impact},  and  \citet{john2019eyeveil} showed that iris recognition accuracy degrades when the eye image is blurred.

\citet{john2019eyeveil} showed that pupil detection rates started to deteriorate with increased Gaussian blur, though their results indicated that iris recognition was maintained over certain ranges of Gaussian blur. Thus, there exists a trade-off between pupil detection and iris identification. Further, providing a detailed (unsmoothed) texture to eye images is vital for most of the deep learning-based architectures such as \cite{park2018learning,park2018deep,yiu2019deepvog}, which learns eye characteristics, such as pupil and iris, based on features from the whole image. Another area of interest for some researchers is tracking the iris textures for gaze estimation, as in \cite{pelzwitzner, Chaudhary_Pelz_2019}. 

\begin{figure}[h]
\begin{center}
\includegraphics[width=0.65\linewidth]{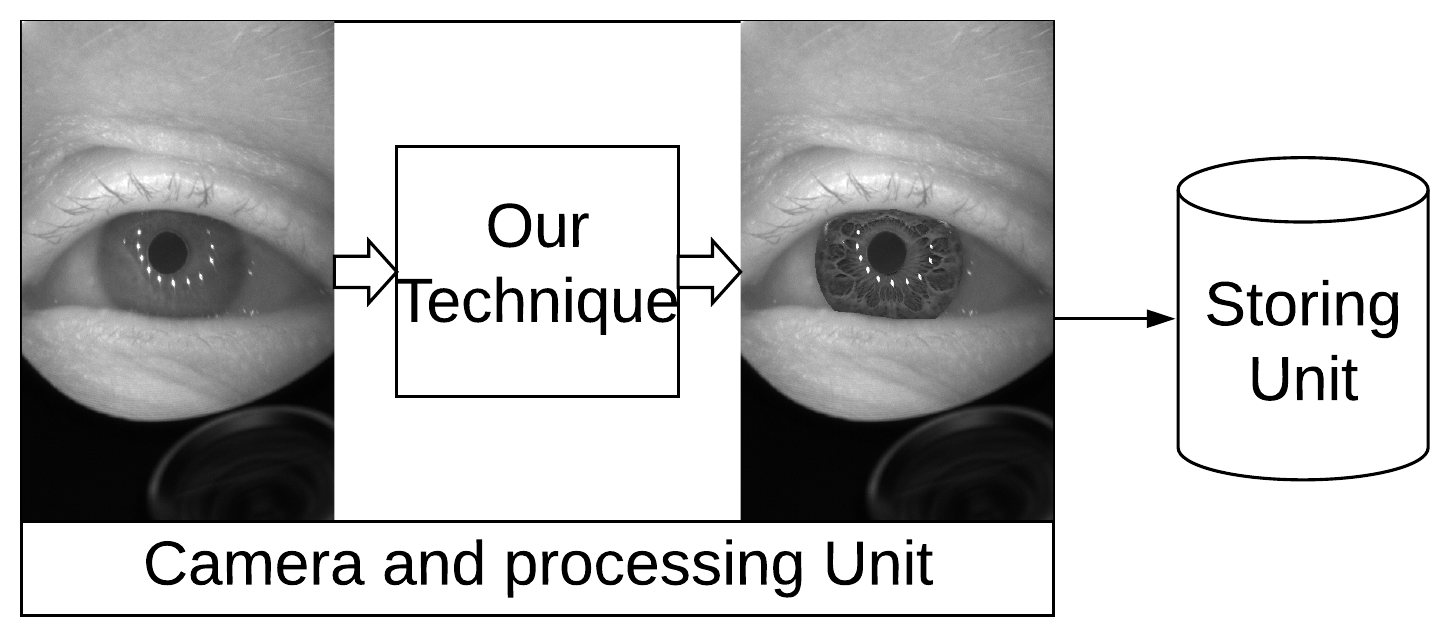}
\end{center}
\caption{Proposed Video Capture pipeline.}
\label{fig:block2}
\end{figure}

The main contribution of this paper is a method to prevent iris identification by a transformation technique based on the rubber sheet model \cite{daugman1993high} which maps every point in the Cartesian coordinate system to a rectangular approximation of its polar coordinates to generate a new uncorrelated iris texture eye image without degrading the estimation of gaze. Figure ~\ref{fig:block2} shows our proposed video capture pipeline.
%The proposed method can be used in the video capture pipeline, as shown in Figure ~\ref{fig:block2}. 

\section{Methods}
\label{methods}
In this section, we discuss the mapping of the identifiable \textit{source} original  image to the non-identifiable \textit{target} iris by the following steps: eye segmentation, iris transformation, and glint replacement. 

\subsection{Eye Segmentation}
Initially, to proceed with a rubber sheet model, we need to segment the region of interest, namely pupil and iris, as we require the pupillary and limbus border points for the rubber sheet model \cite{daugman1993high}. Segmentation algorithms proposed by \cite{vera2019deepeye,kim2019nvgaze,yiu2019deepvog,zhang2015appearance,aayu,garbin2019openeds} have been used to annotate these regions. As our main concern lies in the proof of concept of the iris transformation approach and not on designing an eye segmentation model, we use the pre-trained RITnet model \cite{aayu} as it is capable of isolating the pupil, iris, and sclera regions with high accuracy in real-time. 

\subsection{Iris transformation}
\label{sec:exp}
After the region of the iris and pupil are segmented, we transform the \textit{source} images to \textit{target} iris images. A complex solution is required to replace the iris texture in the given images with a new texture because we must take into account factors such as perspective deformation, pupil dilation, and occlusion with the eyelids. We apply a rubber sheet transformation on the \textit{target} iris with the distance between the iris and pupil boundary being represented by $r$. Note that the rubber sheet model takes into account factors such as pupil dilation and iris deformation \cite{masek2003recognition}%,chen2006localized}.

The goal is to match the \textit{target} iris to the \textit{source} iris, so the iris orientation, position, and size of \textit{target} and \textit{source} should be the same. Thus, after the transformation of the \textit{target} iris to an unwrapped rectangular form, we use cubic interpolation along the $r$ direction to match the maximum \textit{source} iris radius ($R$). To account for iris rotation, we shift the interpolated iris template along the +ve x-direction ($\theta$) based on the elliptical rotation of the source iris. It would be preferable, however, to find the actual iris rotation by incorporating a measure of ocular torsion with the Rubber Sheet model \cite{ong2010measuring,lee2007robust} or by tracking iris texture features \cite{pelzwitzner,Chaudhary_Pelz_2019} in the \textit{source} images, which would support the measurement of torsional eye movements.

%\aayush{based on the elliptical rotation of the source iris as a standalone image dataset was used. However, it is preferred to get exact iris rotation by incorporating the measure of ocular torsion with Rubber Sheet model \cite{ong2010measuring,lee2007robust} or by iris-texture tracking \cite{pelzwitzner,Chaudhary_Pelz_2019} in \textit{source} iris for temporal image-pair as it also allows for the study involving torsional eye movements.}

\begin{figure}
\begin{center}
\includegraphics[width=0.9\linewidth]{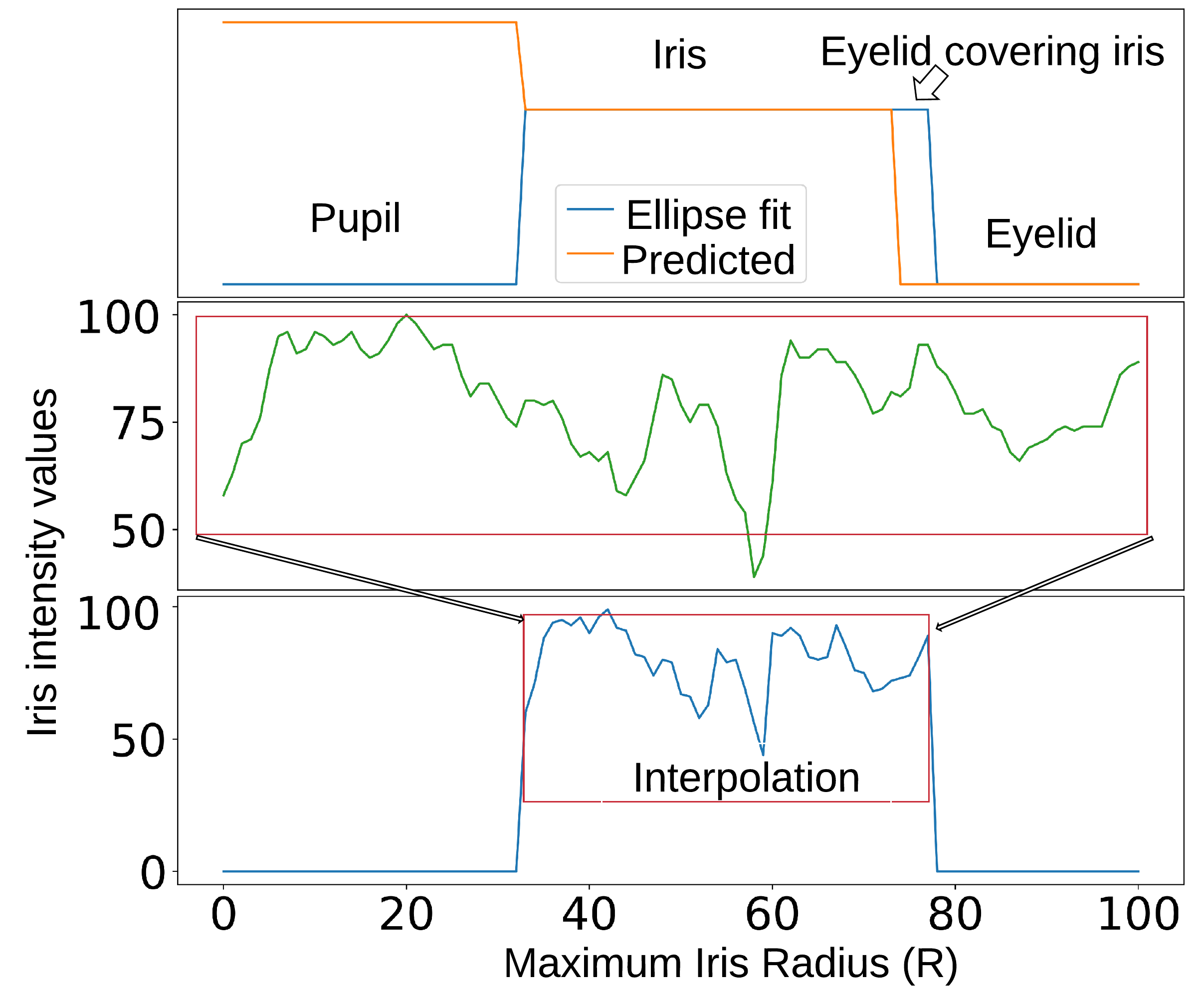}
\end{center}
\caption{Iris texture deformation. (Top) The annotated labels for one radial angle. As we move from the center of the iris to the maximum iris radius ($R$), we encounter pixels representing pupil, iris, and sclera/eyelid regions. Sclera region is also possible as we consider the circle with radius $R$. In this example, the eyelid is covering part of the iris, so we see a gap between the ellipse fit and predicted iris segmented mask. (Middle) The generated texture along that radial angle. (Bottom) After interpolation of the textured pattern, it only covers the desired iris as seen in (Top). Note that the deformation is executed according to the iris ellipse fit, but some iris features are invisible under the eyelids.}
\label{fig:block3}
\end{figure}

Every point on the template generated after rotation must be matched with the \textit{source} iris, taking into account factors such as iris deformation, ellipse shape, non-concentric pupil-iris displacement, and extreme eye positions.
%It is necessary to match every point on the template generated after the rotation with the \textit{source} iris, taking into account factors such as iris deformation, ellipse shape, non-concentric pupil-iris displacement, and extreme eye positions. 
To do this, we use a ratio of the number of pixels of the sclera/eyelid, pupil, and iris in the \textit{source} image along $R$ and fit along the $\theta$ direction through interpolation and matching, as shown in Figure ~\ref{fig:block3}. 

After the properties of iris deformation and eye-region are correctly identified, the rectangular unwrapped region  is converted back into the Cartesian coordinate system with the inverse process of the rubber sheet model. The result is that the derived position of the iris is in the same position as the iris in the \textit{source} image. It is also necessary to match the \textit{source} iris and \textit{generated} iris intensity distribution, which is done by histogram equalization. Finally, each pixel of the iris in the \textit{source} image is replaced by the same pixel in the \textit{generated} iris image to get the final \textit{generated} image. Refer to Figure ~\ref{fig:block} for the basic flow of our methodology.

\subsection{Glints}
The glints were removed from the \textit{target} iris image before the transformations described in section ~\ref{sec:exp}. Because some eye-tracking methodologies rely on glints for gaze estimation, we replace the glints in the same position on the \textit{generated} image. The glints are the brightest region in an image; we threshold these glints based on the digital count on the source image, then replace them on the \textit{generated} image to get our final eye image.

\subsection{Blended Image}
\citet{phillips2011impact} and \citet{john2019eyeveil} convolved the entire eye image with Gaussian kernels to degrade the iris recognition process. Our transformation step does not degrade the eye image; instead, it replaces the iris region with a different high-quality iris texture image even when there is motion blur. We also propose a way to improve on our transformation pipeline by blending the \textit{source} iris with the \textit{generated} iris. The pipeline replaces the region indicated in the blue shaded color box in Figure ~\ref{fig:block}. As we are confident in pupil segmentation, we isolate an elliptical region of 5 pixels around the pupil boundary and find the median digital count. This region in the \textit{generated} iris is replaced by this median digital count allowing pupil detection algorithms to detect the original pupil robustly. Then we blend the \textit{source} iris and the \textit{generated} iris with a weighted elliptical gradient function (Equation~\ref{eq:1}) \cite{hendricks2012rotated} giving high importance to the \textit{generated} iris towards the boundary. These images are referred to as \textit{blended} images in the following sections.
\begin{equation}
w=\frac{((x-h)cos\theta+(y-k)sin\theta)^2}{a^2}+\frac{((y-k)cos\theta-(x-h)sin\theta)^2}{b^2}
\label{eq:1}
\end{equation}
where (h,k,a,b,$\theta$) are standard ellipse parameters.

\section{Dataset and Evaluation} %previously Dataset and Perfomance Evaluation
We evaluated our technique on the publicly available OpenEDS Semantic Segmentation dataset \cite{garbin2019openeds} using the state-of-the-art RITnet model trained on the OpenEDS dataset. Since the ground truth of the OpenEDS test dataset is not publicly available, we compared the performance to the validation set.

To evaluate the performance of the images after alteration to protect privacy, we compared results for the \textit{source}, \textit{generated} and \textit{blended} images based on 1) mean per-class Intersection over Union (mIoU) segmentation results when tested with the trained models; 2) center estimate based on an ellipse fit of the segmentation mask of the pupil; and 3) the reported pupil center based on the open-source software from Pupil Labs \cite{Kassner:2014:POS:2638728.2641695}. For comparison, we also generated another privacy-preserving eye image where every pixel in the iris region was replaced by the median digital count of the iris in that image (referred to as a \textit{median} image). The replacement of the iris region with the median digital count can be useful in any system not relying on the iris features in the image.

\section{Results}
\begin{figure}[h]
\begin{center}
\includegraphics[width=\linewidth]{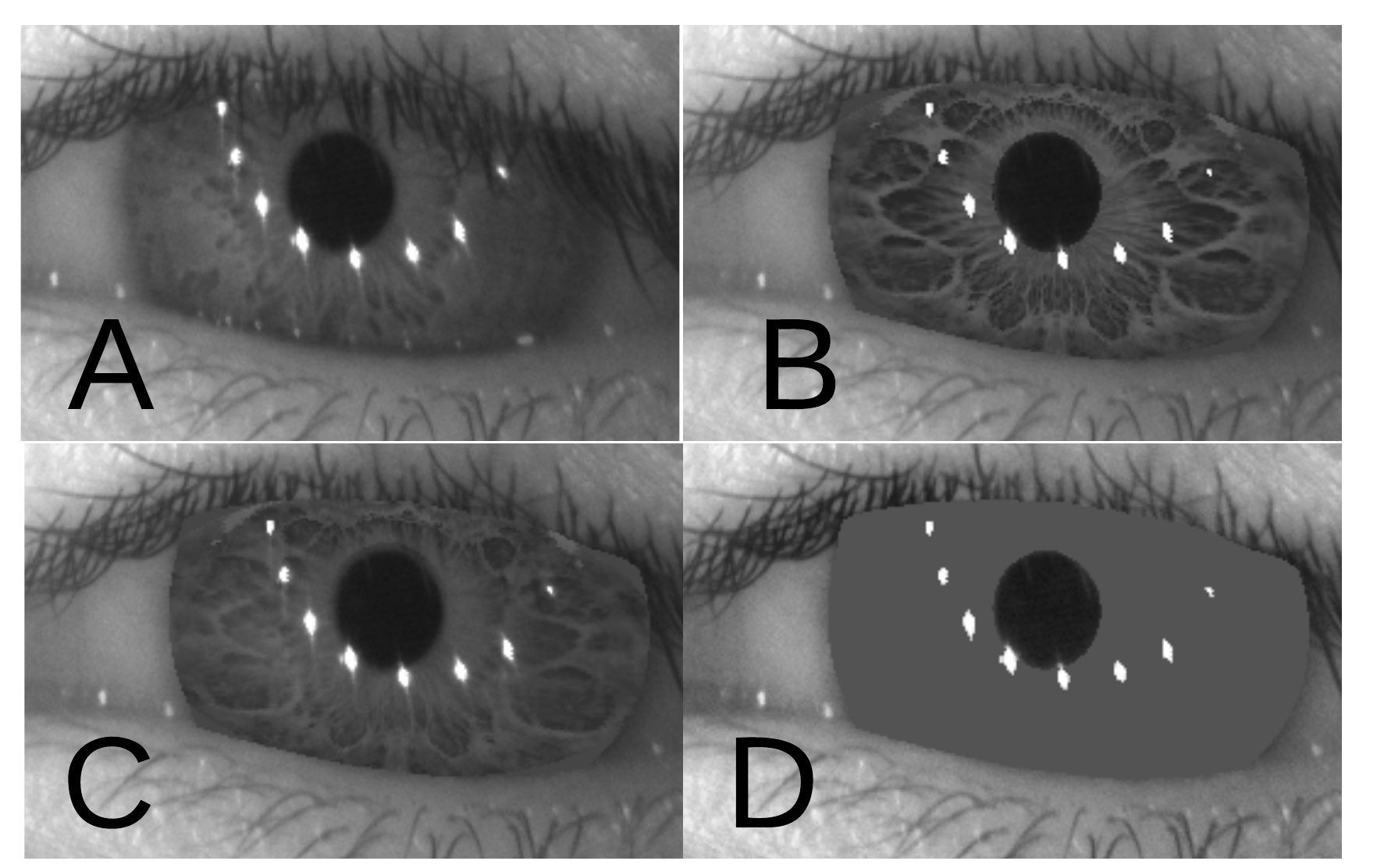}
\end{center}
\caption{Sample Images: (A) Source, (B) Generated, (C) Blended, (D) Median }
%000000265259.png
\label{fig:block4}
\end{figure}
Figure ~\ref{fig:block4} shows an iris-cropped sample of the image from the OpenEDS dataset and its transformation images along with blended and median images. The results are shown in Tables 1-3. Table 1 shows the mIoU for images from the OpenEDS validation set for the \textit{source}, \textit{generated}, \textit{blended}, and \textit{median} image, respectively. There is a small decrement in mIoU performance, never exceeding 2.1\% in the \textit{generated} and \textit{blended} image sets, though the \textit{median} image set (without a replacement iris texture) reaches 2.9\%. The table also shows the per-class accuracies for the pupil, iris, sclera, and background. Table 2 shows the mean square error (MSE) in the estimate of the pupil center based on an ellipse fit on the predicted labels for various images in both horizontal (x) and vertical (y) directions. Note that all the results presented are with respect to the ellipse fit on the ground truth. The $R^2$ value for all the cases was over 0.999.

Table 3 represents the 2D pupil fit results of the video sequence from the images by the open-source Pupil Labs software (V1.8-26). The reported metrics are pupil detection rate, proportion of images with pupil detection confidence over 80\%, and MSE in x and y directions. The MSE in this case was calculated after rejecting 5\% of outliers. Note that the video was not recorded with a Pupil Lab tracker, but the videos were processed keeping all parameters (such as the region of interest, pupil intensity, minimum pupil radius, and maximum pupil radius) constant for all videos. Out of 2403 images, we saw a pupil detection increment in seven \textit{generated} images and 48 \textit{median} images relative to the \textit{source} image, and a decrement in nine \textit{blended} images. Results for samples with pupil detection confidence above 80\% showed an increment of 19, 85 and six samples in \textit{generated},  \textit{median} and \textit{blended} images, respectively. The MSE results were consistent for all the videos.  

\begin{table}[t]
\caption{Comparison of the mean Per class IoU for the original \textit{source}, \textit{generated}, \textit{blended} and \textit{median} images.}
\label{tab:1}  
\begin{tabular}{|l|l|l|l|l|}
\hline
\textbf{Class}      
 &\textbf{Source}&\textbf{Generated}&\textbf{Blended}&\textbf{Median}\\ \hline
mIoU & 95.75 & 93.66  &94.44& 92.85 \\\hline
Pupil& 94.82 & 93.78 & 94.14& 93.85 \\\hline
Iris & 95.61 & 91.29 & 92.97& 90.46 \\\hline
Sclera & 93.11 & 90.53 & 91.40& 87.77 \\\hline
Background & 99.46 & 99.05& 99.25 & 99.31 \\\hline
\end{tabular}
\end{table}

\begin{table}[t]
\caption{MSE in pupil center estimate based on ellipse fit on the segmentation mask with respect to ground truth.}
\label{tab:1}  
\begin{tabular}{|l|l|l|l|l|}
\hline
\textbf{Images} &\textbf{Source}&\textbf{Generated}&\textbf{Blended}&\textbf{Median}\\ \hline
$MSE_x$ & 0.77&0.72 & 0.72 & 1.01\\ \hline
$MSE_y$ & 0.51& 1.40 & 0.88 & 0.98\\ \hline
\end{tabular}
\end{table}
\begin{table}[t]
\caption{Estimates based on Pupil Labs Software \cite{Kassner:2014:POS:2638728.2641695}}
\label{tab:1}  
\begin{tabular}{|l|l|l|l|l|}
\hline
\textbf{Images} &\textbf{Source}&\textbf{Generated}&\textbf{Blended}&\textbf{Median}\\ \hline 
Detection Rate & 76.70\% & 76.99\%& 76.36\%& 78.73\%\\ \hline
$>80\%$ confidence & 68.29\%& 69.12\%& 68.58\%& 71.83\%\\ \hline
$MSE_x$& N/A & 0.08 & 0.06 & 0.08\\ \hline
$MSE_y$ & N/A & 0.25 & 0.09 & 0.23\\ \hline
\end{tabular}
\end{table}

\section{Discussion}
Current video-based systems allow individuals to be identified through their eye videos, which is a privacy concern. We present an approach to preserve privacy by replacing the iris with uncorrelated iris texture maps (Hamming Distance (HD)= 0.47 $\pm$ 0.01 with its \textit{source} iris based on encoding procedure of \citet{masek2003recognition} and matching technique of \citet{daugman2009iris}). The iris texture is replaced by taking advantage of the Daugman's rubber sheet model and RITnet. The proposed method handles iris deformation due to perspective projection, specular glints, and elliptical-shaped iris/pupil transformation. We show that this transformation degrades the performance by 2.09\% in mIoU, 1.04\% in pupil segmentation, and ~0.47 pixels root MSE. Instead of an uncorrelated image, the introduction of the weighted elliptical gradient blending (HD= 0.27 $\pm$ 0.04) with the \textit{source} image only degraded the performance by 1.31\%, 0.68\%, ~0.23 pixels, respectively.

%Current video-based systems allow individuals to be identified through their eye videos, which is a privacy concern. We present an approach to preserve privacy based on uncorrelated \aayush{(Hamming Distance (HD)= 0.47 $\pm$ 0.01 with its \textit{source} iris based on encoding procedure of \citet{masek2003recognition} and matching technique of \citet{daugman2009iris})} iris texture maps taking advantage of the rubber sheet model and RITnet. The proposed method handles iris deformation due to perspective projection, specular glints, and elliptical-shaped iris/pupil transformation. We show that this transformation degrades the performance by 2.09\% in mIoU, 1.04\% in pupil segmentation, and ~0.47 pixels root MSE. Instead of an uncorrelated image, the introduction of the weighted elliptical gradient blending \aayush{(HD= 0.27 $\pm$ 0.04)} with the \textit{source} image improved the performance to a decrement of 1.31\%, 0.68\%, ~0.23 pixels, respectively.

We also show similar results when the videos are fed to the open-source Pupil Labs software. There is a boost of over 2\% in the pupil detection and proportion of the confidence over 80\% metric in the \textit{median} image since it is easy for the algorithm to detect the clean pupil/iris boundary. This suggests that any eye-tracking methodology focused on detecting pupil based on edges can simply use the \textit{median} image. However, the disadvantage of the use of the \textit{median} iris image is a significant decrement in neural network-based performance in the segmentation task. We can argue that the results for the segmentation can be improved if such a network is trained on \textit{generated}, \textit{blended}, and \textit{median} images, but the primary concern of this paper lies in the performance comparison without re-training the architectures. Overall, our results for all the cases are comparable to real video results (\textit{source}) and shows the benefit of using this approach instead of degrading eye videos by blurring.

The proposed approach handles the image generation with artificial/real eye images. With the evolution of deep learning architectures, especially generative adversarial networks (GANs) \cite{goodfellow2014generative}, there is the possibility of creating unidentifiable videos based on learned features. However, the more significant challenges for such architectures are the requirement of huge databases with proper ground truth, no proper validation metric, and no study of gaze estimation on GAN generated images.% Additionally, our setup can be used in order to generate the ground truth for such cases.

Our method improves on previous attempts to limit identification through iris recognition in eye-tracking videos. However, the study has a number of limitations. First, a person can be identified by eye features other than the iris, such as the sclera, eye corners, eyelids, facial structure, and even eye movements. Our current study does not account for such factors. Since we also segment the sclera, a similar sclera-generation technique could be implemented in our pipeline. Secondly, we have tried to adapt the lighting condition of the iris-based histogram equalization, but it does not account for positioning and occlusion of the IRLEDs. Note that histogram equalization (as in Figure ~\ref{fig:block}) boosted the performance by ~2\%. A more general technique to match lighting and other performance would be valuable for our model. Furthermore, the replacement of a part of one image with another can have unintended consequences. One example of this is when was features like eyelashes covering the iris are replaced by the generated iris without the eyelashes, leading to discontinuities in the iris/pupil border. In the future, we plan to incorporate GANs in order to generate more realistic iris textures that can simulate other textures. In the future we plan to study the processing load of pre-processing steps, light-weight architecture \cite{aayu}, rubber sheet mapping, and its inverse with possible vectorization techniques to support real-time use.

\section{Conclusion}
We have proposed a new method for altering eye videos in a manner that preserves observer privacy without significantly affecting the accuracy of feature-based and appearance-based gaze-tracking methods. The new images have no correlation with the original video in terms of iris patterns, which prevents iris recognition while permitting pupil, corneal-reflection, and iris-texture tracking.

\bibliographystyle{ACM-Reference-Format}
\bibliography{ref}

\end{document}